# Formation Maneuvering Control of Multiple Nonholonomic Robotic Vehicles: Theory and Experimentation

Milad Khaledyan and Marcio de Queiroz
Department of Mechanical & Industiral Engineering
Louisiana State University
Baton Rouge, La 70803

June 22, 2017

### Abstract

In this paper, we present a new leader-follower type solution to the formation maneuvering problem for multiple, nonholonomic wheeled mobile robots. The solution is based on the graph that models the coordination among the robots being a spanning tree. Our decentralized control law ensures, in the least squares sense, that the robots globally acquire a given planar formation while the formation as a whole globally tracks a desired trajectory. The control law is first designed at the kinematic level and then extended to the dynamic level. In the latter, we consider that parametric uncertainty exists in the equations of motion. These uncertainties are accounted for by employing an adaptive control scheme. The proposed formation maneuvering controls are demonstrated experimentally and numerically.

## 1   Introduction

The field of multi-agent systems has attracted the attention of many systems and control researchers for the past decade. The main interest is in studying the coordinated behavior of multiple robots as they perform tasks unsuitable and/or too complex for a single robot. Formation control is a type of coordinated behavior where mobile agents are required to autonomously configure into a desired spatial pattern. Formation maneuvering refers to the special case where the desired formation is not static, but moves in space as a virtual rigid body according to a pre-defined trajectory.

Most formation control results are based on point-mass type models for the agent's motion, such as the single- and double-integrator models. For example, see (4; 14; 23; 36; 40) for single-integrator results and (3; 6; 30; 31) for double-integrator results. On the other hand, some results have used more sophisticated models that account for the agent kinematics/dynamics. One of two models are used in these cases: the fully-actuated (holonomic) Euler-Lagrange model, which includes robot manipulators, spacecraft, and some omnidirectional mobile robots; or the nonholonomic (underactuated) model, which accounts for velocity constraints that typically occur in the vehicle motion (e.g., differentially-driven wheeled mobile robots and air vehicles). In the nonholonomic case, models can be further subdivided into two categories: the purely kinematic model where the control inputs are at the velocity level, and the dynamic model where the inputs are at the actuator level. Examples of work based on the Euler-Lagrange model include (5; 7; 8; 11; 20; 26; 32; 38; 39). Formation control results based on nonholonomic kinematic models can be found in (1; 21; 28; 29; 35). Designs for nonholonomic dynamic models appeared in (9; 12; 13; 15; 27; 42).

This paper is concerned with the formation maneuvering problem for unicycle-type nonholonomic robotic vehicles with dynamics. Our intent is to contribute a new control solution to this problem. To this end, we consider a leader-follower type coordination scheme, composed of a primary leader, secondary leaders, and followers, where the inter-vehicle interactions are modeled by a spanning tree graph. This scheme motivates us to introduce two categories of position errors in the control design: individual tracking errors (relative to the desired trajectory of each vehicle) and coordination errors for leader-follower pairs in the spanning tree. The latter category is inspired by the error used in (21; 35; 37), and serves the purpose of coupling the motion of the individual vehicles in the formation. As a result, only the global position of the primary leader is required along with the relative position of the robots connected in the graph. Based on a composite Lyapunov analysis of the primary leader's tracking error and all coordination errors, we construct a decentralized, nonlinear control law that ensures all the errors are globally convergent to zero (and therefore, the formation maneuvering problem is solved) in the least squares sense (see Section 5 for details). This is accomplished by exploiting the pentadiagonal structure of a matrix in the Lyapunov analysis. For ease of explanation, the proposed control law is first developed for the nonholonomic kinematic model, and then extended to account for the vehicle dynamics using the backstepping technique (24). In this extension, we assume the parameters in the dynamic equations are unknown, and design an adaptive controller to account for this uncertainty.

Despite similarities in the open-loop error dynamics, our result differs from (21; 35) in several aspects: i) we use a spanning tree for the coordination graph to minimize the number of control links; ii) we employ a different Lyapunov function candidate and, as a result, a different control law; iii) the stability analysis is simpler since we are only concerned with the negative definiteness of the Lyapunov function derivative in the least squares sense; and finally, iv) we account for the vehicle dynamics.

The paper is organized as follows. In Section 2, we review some concepts of graph theory that will be used in the formation control design. The equations of motion for the nonholonomic robots are introduced in Section 3. This is followed by a formal statement of the formation maneuvering problem in Section 4. The control law for the kinematic equations are given in 5 and extended to the uncertain dynamic equations in 6. Experimental results for the kinematic-level controller and simulation results for the dynamics-level controller are presented in Section 7. The paper concludes with a summary of the results in Section 8. A preliminary version of the present paper appeared in (18), where only the kinematic control problem was solved and no experimental results were included.

## 2    Graph Theory Concepts

This paper utilizes some concepts of graph theory, which are reviewed below (16).

- An *undirected graph* $G$ is a pair $(V, E)$ where $V = \{1, ..., n\}$ is the set of vertices and $E \subset V \times V$ is the set of undirected edges that connect two different vertices, i.e., if vertex pair $(i, j) \in E$ then so is $(j, i)$.

- A *path* is a trail that goes from an origin vertex to a destination vertex by traversing edges of the graph.

- An undirected graph is *connected* if there is a path between every pair of vertices of $G$.

- A *tree* is a connected graph in which two vertices are connected exactly by one path. Note that in a tree, cycles cannot exist.

- A *spanning tree* for a connected graph $G$ is a tree containing all the vertices of $G$. In other words, a *spanning tree* of a graph of $n$ vertices is a subset of $n-1$ edges that form a tree.

## 3    System Model

We consider a system of $n$ wheeled nonholonomic robots moving autonomously on the plane. We assume the wheels operate under the conditions of pure rolling and no slipping. The $i$th robot is depicted in Figure 1, where $\{X_0, Y_0\}$ is a reference frame fixed to the Earth, and $\{X_i, Y_i\}$ is the moving reference frame attached to the robot such that the $X_i$-axis is aligned with robot's heading direction, which is denoted by the angle $\theta_i$ and measured counterclockwise from the $X_0$-axis. It is assumed that the robot's center of mass is coincident with its center of rotation located at point $C_i$.

We consider that each robot is a unicycle vehicle governed by following equations of motion (10)

$$\dot{q}_i = S(\theta_i)\eta_i \tag{1a}$$

$$M_i\dot{\eta}_i + D_i\eta_i = u_i \tag{1b}$$

for $i = 1, 2, ..., n$. The robot kinematics is given by (1a), where $q_i = (x_i, y_i, \theta_i)$

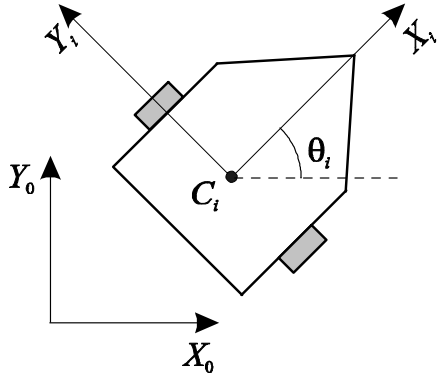

Figure 1: Schematic of the unicycle vehicle.

denotes the position and orientation of $\{X_i, Y_i\}$ relative to $\{X_0, Y_0\}$, $\eta_i = (v_i, \omega_i)$, $v_i$ is the robot's translational speed in the direction of $\theta_i$, $\omega_i$ is the robot's angular speed about the vertical axis passing through $C_i$, and

$$S(\theta_i) = \begin{bmatrix} \cos \theta_i & 0 \\ \sin \theta_i & 0 \\ 0 & 1 \end{bmatrix}. \tag{2}$$

The robot dynamics is described by (1b), where $M_i = \mathrm{diag}(m_i, J_i)$, $m_i$ and $J_i$ are the vehicle mass and moment of inertia about the vertical axis, respectively, $D_i \in \mathbb{R}^{2 \times 2}$ is the constant damping matrix, and $u_i \in \mathbb{R}^2$ represents the force/torque-level control input provided by the drive train actuators.

## 4 Problem Statement

Our control objective is to ensure that the system of $n$ nonholonomic robots acquires a specified two-dimensional formation and maneuvers cohesively according to a predefined desired trajectory.

The information exchange between robots is modeled as a spanning tree described by the graph $G(V, E)$. The spanning tree guarantees that the least number of control links (edges) between the robots in the graph is used to meet the control objective. The spanning tree is also motivated by our proposed collaboration protocol for the robots, which includes three categories of agents. In particular, we will designate one robot as the *primary leader*, which for notational convenience will be robot 1, who knows its global position (i.e., with respect to $\{X_0, Y_0\}$). The *secondary leaders* are robots who have a dual role: they follow the primary leader or a secondary leader, but also serve as a leader for other robots. Finally, the *followers* are robots that simply follow the primary or a secondary leader with no leadership role. Secondary leaders and followers can only measure their *relative* position to other robots to which they are connected in $G$.

**Remark 1** *The control design of this paper also applies to any connected graph. Specifically, one can introduce extra edges in the spanning tree to add redundancy and robustness to the system in case of robot failure.*

We will quantify the control objective by defining two types of position errors. The first one is the standard *tracking* error for each individual robot:

$$e_i = q_{di} - q_i, \quad i \in V \tag{3}$$

where $q_{di}(t) \in \mathbb{R}^3$ denotes the desired trajectory for robot $i$ (below we describe how $q_{di}$ is generated) dictated by the desired formation shape and maneuver. We also introduce the following *coordination* error between robots $i$ and $j$ (21; 35)

$$\varepsilon_{ij} = e_i - e_j, \quad (i, j) \in E \tag{4}$$

where $E$ is edge set of the spanning tree. The coordination errors will couple the motion of robots so that the desired formation maneuvering is accomplished by only measuring $q_1$ and $q_i - q_j$, $(i, j) \in E$. That is, from (3) and (4), we have that

$$\varepsilon_{ij} = q_{di} - q_{dj} - (q_i - q_j), \tag{5}$$

which is only dependent on the relative position of robots $i$ and $j$. Our problem statement is then to design $u_i = u_i(t, \theta_i, \eta_i, q_1, q_i - q_j)$, $\forall i \in V$ and $\forall (i, j) \in E$ in (1) such that

$$(e_i(t), \varepsilon_{ij}(t)) \to 0 \quad as \ t \to \infty \tag{6}$$

for $\forall i \in V$ and $\forall (i, j) \in E$.

The desired trajectory of each robot needs to satisfy the nonholonomic constraint of the unicycle vehicle given by (1a). That is,

$$\dot{q}_{di} = S(\theta_{di}) \eta_{di} \tag{7}$$

where $q_{di} = (x_{di}, y_{di}, \theta_{di})$, and $\eta_{di} = (v_{di}, \omega_{di})$ includes the desired translational and rotational speeds such that $v_{di}(t) \neq 0$, $\forall t$. This condition on $v_{di}$ stems from the potential singularity in the relationship $\omega_{di} = (\dot{x}_{di}\ddot{y}_{di} - \ddot{x}_{di}\dot{y}_{di})/v_{di}^2$, which can be derived from (7) and its derivative. We assume the desired trajectory is generated such that $q_{di}(t)$, $\dot{q}_{di}(t)$, $\eta_{di}(t)$, and $\dot{\eta}_{di}(t)$, $i = 1, \ldots, n$ are bounded for all time. The signal $q_{di}(t)$ is usually known in advance and therefore can be stored on the $i$th robot's microcontroller.

**Remark 2** *Measurement of the relative linear positions in (5) can accomplished using, for example, a rangefinder, camera, and compass. The global position of the primary leader can be measured using an odometer coupled to a Kalman filter. The robots' heading angles can be measured with a compass, while velocity signals can be determined by numerically differentiating and low-pass filtering the position signals. That is, all measurements needed to implement the control* law can be done via onboard sensors and broadcast between robots using a local wireless network. In other words, no external agent or central command is necessary for implementation of the proposed formation control system.

**Remark 3** *In theory, a formation control law could be designed just based on (3) such that each robot tracks its own desired trajectory, independently of one another. This however has the disadvantage that a) perturbations from the desired trajectory will distort the formation shape without a mechanism for self-correction, and b) the control will depend on the global position of all robots. These problems are avoided by introducing (4) in the control scheme.*

# 5   Kinematic Control

We first consider the kinematic control problem by treating $\eta_i$ in (1a) as the control input. To this end, we begin by expressing the tracking error (3) with respect to the moving reference frame $\{X_i, Y_i\}$ as follows (10; 17)

$$s_i = R^{\mathsf{T}}(\theta_i)e_i \tag{8}$$

where $R(\theta_i)$ is the rotation matrix defined as

$$R(\theta_i) = \begin{bmatrix} \cos\theta_i & -\sin\theta_i & 0 \\ \sin\theta_i & \cos\theta_i & 0 \\ 0 & 0 & 1 \end{bmatrix}. \tag{9}$$

Differentiating (8) with respect to time yields

$$\begin{aligned} \dot{s}_i &= \dot{R}^{\mathsf{T}}(\theta_i)e_i + R^{\mathsf{T}}(\theta_i)\left(\dot{q}_{di} - \dot{q}_i\right) \\ &= \omega_i Q s_i + R^{\mathsf{T}}(\theta_i)S(\theta_{di})\eta_{di} - P\eta_i \end{aligned} \tag{10}$$

where

$$Q = \begin{bmatrix} 0 & 1 & 0 \\ -1 & 0 & 0 \\ 0 & 0 & 0 \end{bmatrix}, \quad P = \begin{bmatrix} 1 & 0 \\ 0 & 0 \\ 0 & 1 \end{bmatrix}, \tag{11}$$

$\dot{R}^{\mathsf{T}}(\theta_i) = \omega_i Q R^{\mathsf{T}}(\theta_i)$, and $S(\cdot)$ was defined in (2).

Now, consider the Lyapunov function candidate

$$V(z) = \frac{1}{2}s_1^T s_1 + \frac{1}{2}\sum_{(i,j)\in E}\varepsilon_{ij}^{\mathsf{T}}\varepsilon_{ij} = \frac{1}{2}z^{\mathsf{T}}z \tag{12}$$

where

$$z = (s_1, \ldots, \varepsilon_{ij}, \ldots) \in \mathbb{R}^{3n}, \quad \forall(i,j) \in E. \tag{13}$$

It is important to note that (13) is a function of all coordination errors but only the primary leader's tracking error.

The time derivative of (12) is given by

$$
\begin{aligned}
\dot{V} = \ & s_1^\mathsf{T} \left[ R^\mathsf{T}(\theta_1) S(\theta_{d1}) \eta_{d1} - P \eta_1 \right] + \sum_{(i,j) \in E} \varepsilon_{ij}^\mathsf{T} \left[ S(\theta_{di}) \eta_{di} \right. \\
& \left. - S(\theta_i) \eta_i - S(\theta_{dj}) \eta_{dj} + S(\theta_j) \eta_j \right]
\end{aligned}
\tag{14}
$$

where we used the fact that $Q$ is skew-symmetric. After some manipulations, we can rewrite (14) in the following form

$$
\dot{V} = z^\mathsf{T} \left[ K(\theta) \eta + H(\theta_1, t) \right]
\tag{15}
$$

where $\eta = (\eta_1, \ldots, \eta_n) \in \mathbb{R}^{2n}$, $\theta = (\theta_1, \ldots, \theta_n) \in \mathbb{R}^n$, and $K \in \mathbb{R}^{3n \times 2n}$ and $H \in \mathbb{R}^{3n}$ are defined as

$$
K(\theta) = \begin{bmatrix}
-P & 0_{3\times2} & 0_{3\times2} & 0_{3\times2} & \cdots & 0_{3\times2} \\
-S(\theta_1) & S(\theta_2) & 0_{3\times2} & 0_{3\times2} & \cdots & 0_{3\times2} \\
0_{3\times2} & -S(\theta_2) & S(\theta_3) & 0_{3\times2} & \cdots & 0_{3\times2} \\
0_{3\times2} & 0_{3\times2} & -S(\theta_3) & S(\theta_4) & 0_{3\times2} & 0_{3\times2} \\
\ddots & \ddots & \ddots & \ddots & \ddots & \ddots \\
0_{3\times2} & 0_{3\times2} & \cdots & 0_{3\times2} & -S(\theta_{n-1}) & S(\theta_n)
\end{bmatrix}
\tag{16}
$$

$$
H(\theta_1, t) = \begin{bmatrix}
R^\mathsf{T}(\theta_1) S(\theta_{d1}) \eta_{d1} \\
S(\theta_{d1}) \eta_{d1} - S(\theta_{d2}) \eta_{d2} \\
\vdots \\
S(\theta_{d(n-1)}) \eta_{d(n-1)} - S(\theta_{dn}) \eta_{dn}
\end{bmatrix}.
\tag{17}
$$

In the following theorem, we give a solution to the kinematic control problem that is valid in the *least squares sense*.

**Theorem 4** The kinematic control

$$
\eta = -K^\dagger(\theta) \left[ \lambda_1 z + H(\theta_1, t) \right],
\tag{18}
$$

where $\lambda_1$ is a $3n \times 3n$ diagonal, positive-definite, control gain matrix and $K^\dagger$ is the *pseudo-inverse* of (16) defined as

$$
K^\dagger = (K^\mathsf{T} K)^{-1} K^\mathsf{T},
\tag{19}
$$

ensures that (6) is met with global exponential decay and that all other system signals are globally bounded.

**Proof.** If we can specify $\eta$ to satisfy

$$
K \eta + H = -\lambda_1 z,
\tag{20}
$$

then we know from (15) that

$$
\dot{V} = -z^\mathsf{T} \lambda_1 z.
\tag{21}
$$

Along with the form of (12), this indicates that $z = 0$ is exponentially stable (19).

Notice that $K$ in (16) does not have full row rank since the system (20) is overdetermined. Therefore, the exact solution

$$\eta = -K^{\mathsf{T}}(KK^{\mathsf{T}})^{-1}(\lambda_1 z + H) \qquad (22)$$

cannot be used. However, the fact that $K$ has full column rank means the matrix $K^{\mathsf{T}}K$ is invertible (see Appendix A for proof). Therefore, we seek a solution for (20) that minimizes the energy of the error (25)

$$J(\eta) = \|(-\lambda_1 z - H) - K\eta\|^2 \qquad (23)$$

where $\|\cdot\|$ denotes the Euclidean norm. This solution, which we call the *least squares solution*, is given by (18). Therefore, we say that (21) and the global exponential stability of $z = 0$ hold in the least squares sense under the kinematic control law (18).

From the definition of $z$, we know that $s_1(t), \varepsilon_{ij}(t) \to 0$ as $t \to \infty$, $(i, j) \in E$ globally exponentially fast. From (8), we then know that $e_1(t) \to 0$ as $t \to \infty$ globally exponentially fast. Given the spanning tree nature of the graph, we can recursively use (4) starting from the primary leader to show that $e_i(t) \to 0$ as $t \to \infty$, $i = 2, \ldots, n$ globally exponentially fast.

Based on (3) and the assumption that $q_{di}(t)$, $i = 1, \ldots, n$ is bounded, we know that $q_i(t)$, $i = 1, \ldots, n$ is bounded for all time. Since $e_i(t)$, $i = 1, \ldots, n$ is bounded, we know from (8) that $s_i(t)$, $i = 2, \ldots, n$ is bounded. From (16) and (17), we have that $K(\theta)$ and $H(\theta_1, t)$ are bounded. Thus, we know from (18) that $\eta(t)$ is bounded. From (1a) and (2), we can conclude that $\dot{q}_i(t)$, $i = 1, \ldots, n$ is bounded. ∎

# 6 Adaptive Dynamics Control

In this section, we extend the kinematic control law to account for the robot dynamics (1b) so we can design actuator-level control inputs. We will assume that uncertainties exist in the parameters of (1b), which will be accounted for by the design of an adaptive control law. To this end, we exploit the fact that (1b) is linear in the unknown parameters:

$$M_i \mu + D_i \eta_i = Y_i(\mu, \eta_i)\phi_i, \quad \forall \mu = (\mu_1, \mu_2) \qquad (24a)$$

where

$$Y_i = \begin{bmatrix} \mu_1 & 0 & v_i & \omega_i & 0 & 0 \\ 0 & \mu_2 & 0 & 0 & v_i & \omega_i \end{bmatrix} \qquad (25)$$

is the $2 \times 6$ regression matrix,

$$\phi_i = (m_i, J_i, [D_i]_{11}, [D_i]_{12}, [D_i]_{21}, [D_i]_{22}) \qquad (26)$$

is the *unknown*, constant parameter vector, and $[\cdot]_{ij}$ denotes the $ij$th element of the matrix.

The formation controller will include a dynamic estimate of (26), whose adaptation law will be part of the control design. To this end, let $\hat{\phi}_i(t) \in \mathbb{R}^6$ be the parameter estimate for the $i$th robot, and define the corresponding parameter estimation error as

$$\tilde{\phi}_i = \hat{\phi}_i - \phi_i. \tag{27}$$

We also introduce the new error variable

$$\sigma = \eta - \eta_f, \tag{28}$$

where $\eta_f \in \mathbb{R}^{2n}$ denotes the *fictitious* velocity input, to facilitate the use of the backstepping technique (24). Finally, we introduce the new Lyapunov function candidate

$$V_a = V(z) + \frac{1}{2}\sigma^\mathsf{T} M \sigma + \frac{1}{2}\tilde{\phi}^\mathsf{T}\Gamma^{-1}\tilde{\phi} \tag{29}$$

where $V$ was defined in (12), $M = \operatorname{diag}(M_1, ..., M_n)$, $\tilde{\phi} = (\tilde{\phi}_1, ..., \tilde{\phi}_n) \in \mathbb{R}^{6n}$, and $\Gamma \in \mathbb{R}^{6n \times 6n}$ is constant, diagonal, and positive definite.

The following theorem depicts the main result of the paper.

**Theorem 5** The adaptive control

$$u = -\lambda_2 \sigma - K^\mathsf{T}(\theta)z + Y(\dot{\eta}_f, \eta)\hat{\phi} \tag{30}$$

$$\eta_f = -K^\dagger(\theta)\left[\lambda_1 z + H(\theta_1, t)\right] \tag{31}$$

$$\dot{\hat{\phi}} = -\Gamma Y^\mathsf{T}(\dot{\eta}_f, \eta)\sigma, \tag{32}$$

where $u = (u_1, ..., u_n) \in \mathbb{R}^{2n}$, $\hat{\phi} = (\hat{\phi}_1, ..., \hat{\phi}_n)$, $\lambda_1$ is a $3n \times 3n$ diagonal, positive-definite, control gain matrix, and $\lambda_2$ is a $2n \times 2n$ diagonal, positive-definite, control gain matrix, ensures that (6) is satisfied in the least squares sense with global asymptotic decay and that all other system signals are globally bounded.

**Proof.** After taking the time derivative of (29), we obtain

$$\dot{V}_a = z^\mathsf{T}\left[K(\theta)\eta + H(\theta_1, t)\right] + \sigma^\mathsf{T}\left(u - D\eta - M\dot{\eta}_f\right) + \tilde{\phi}^\mathsf{T}\Gamma^{-1}\dot{\tilde{\phi}}$$

$$= z^\mathsf{T}\left[K(\theta)\eta_f + H(\theta_1, t)\right] + \sigma^\mathsf{T}\left[u - Y_i(\dot{\eta}_f, \eta)\phi_i + K^\mathsf{T}(\theta)z\right] + \tilde{\phi}^\mathsf{T}\Gamma^{-1}\dot{\tilde{\phi}} \tag{33}$$

where (1b), (15), (24a), and (28) were used, $D = \operatorname{diag}(D_1, ..., D_n)$, and $Y(\dot{\eta}_f, \eta) = \operatorname{diag}(Y_1(\dot{\eta}_{f_1}, \eta_1), ..., Y_n(\dot{\eta}_{f_n}, \eta_n))$. Now, substituting (30)-(32) into (33) yields

$$\dot{V}_a = -z^\mathsf{T}\lambda_1 z - \sigma^\mathsf{T}\lambda_2 \sigma. \tag{34}$$

From (29) and (34), we know that $\left(z(t), \sigma(t), \tilde{\phi}(t)\right)$ are bounded. We can also invoke the LaSalle-Yoshizawa Theorem to conclude that $(z(t), \sigma(t)) \to 0$ as $t \to \infty$. Therefore, we can follow the arguments used in the proof of Theorem 1 to show that (6) is satisfied globally and asymptotically. Since $\tilde{\phi}(t)$ is bounded, we know from (27) that $\dot{\tilde{\phi}}(t)$ is bounded.

Since $q_{di}(t)$ is bounded by assumption, we know $q_i(t)$ is bounded for all time from (3). Again, we can follow the proof of Theorem 1 to show that $s_i(t)$ and $\eta_f(t)$ are bounded. Then, using (28), we have that $\eta(t)$ is bounded. From (1a) and (2), we conclude that $\dot{q}_i(t)$ is bounded. As shown in Appendix B, $\dot{\eta}_f$ is a function of the variables $\theta$, $\eta$, $z$, and $t$; hence, we know that $\dot{\eta}_f(t)$ is bounded. Consequently, we can show $u(t)$ and $\dot{\tilde{\phi}}(t)$ are bounded using (30) and (32). Finally, we can use (1b) to conclude that $\dot{\eta}_i(t)$ is bounded.  ∎

**Remark 6** *Note that (31) is the right-hand side of (18). Also, the expression for its derivative $\dot{\eta}_f$, which is needed in (30) and (32), is explicitly given in Appendix B.*

# 7  Control Evaluation

## 7.1  Kinematic Control Experiment

To demonstrate the performance of the kinematic controller from Section 5, we conducted an experiment on the *Robotarium* system (34), which is a swarm robotics testbed that uses the GRITSBot as the mobile robot platform (33). The testbed arena has a $130 \times 180$ cm$^2$ area on which multiple robots can be deployed. The GRITSBot is a low-cost, wheeled robot equipped with a suite of onboard sensors, wireless communication, battery, and processing boards, and has a footprint of approximately $3 \times 3$ cm$^2$. An overhead camera and a unique identification tag atop each robot's chassis provide a position tracking system for their motion. A picture of Robotarium and the GRITSBot are shown in Figure 2. Robotarium is ideal for testing kinematic control laws since it uses velocity-level commands as inputs to the robots with the low-level, velocity control loop being invisible to the user.

The desired formation maneuver was a regular pentagon that moves as a virtual rigid body around a circle. To this end, the desired trajectory for the geometric center of the pentagon was chosen as $q_d^{cg}(t) = (5\cos t \text{ cm}, 5\sin t \text{ cm}, t + \pi/2 \text{ rad})$. From this trajectory, we determined the corresponding trajectory for each robot to be governed by (7) with $\eta_{di} = (5 \text{ cm/s}, 1 \text{ rad/s})$, $i = 1, ..., n$, and initial conditions

$$
\begin{aligned}
q_{d1}(0) &= (5 \text{ cm}, 10 \text{ cm}, \pi/2 \text{ rad}), \\
q_{d2}(0) &= (5 + 10\cos \pi/10 \text{ cm}, 10\sin \pi/10 \text{ cm}, \pi/2 \text{ rad}), \\
q_{d3}(0) &= (5 + 10\sin \pi/5 \text{ cm}, -10\cos \pi/5 \text{ cm}, \pi/2 \text{ rad}), \\
q_{d4}(0) &= (5 - 10\sin \pi/5 \text{ cm}, -10\cos \pi/5 \text{ cm}, \pi/2 \text{ rad}), \\
q_{d5}(0) &= (5 - 10\cos \pi/10 \text{ cm}, 10\sin \pi/10 \text{ cm}, \pi/2 \text{ rad}).
\end{aligned}
$$

The desired formation at $t = 0$ along with the desired trajectory for the geometric center of the pentagon is shown in the Figure 3. The graph was selected as the spanning tree shown in Figure 4 with edge set $E = \{(1, 2), (2, 3), (3, 4), (4, 5)\}$. The initial positions and orientations of the robots were randomly set to

$$
\begin{aligned}
q_1(0) &= (2.37 \text{ cm}, 8 \text{ cm}, 0.0162 \text{ rad}), \\
q_2(0) &= (17.5 \text{ cm}, 6 \text{ cm}, 0.0218 \text{ rad}), \\
q_3(0) &= (2.06 \text{ cm}, -1.36 \text{ cm}, -0.0031 \text{ rad}), \\
q_4(0) &= (-9.9 \text{ cm}, -11.49 \text{ cm}, 0.0517 \text{ rad}), \\
q_5(0) &= (-4.45 \text{ cm}, 8.62 \text{ cm}, -0.0452 \text{ rad}).
\end{aligned}
$$

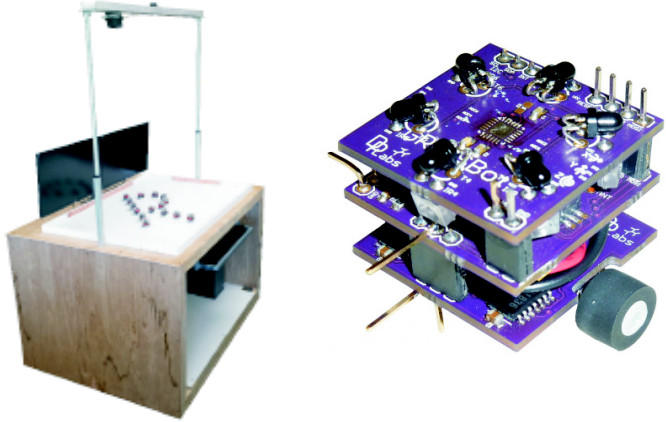

Figure 2: The Robotarium (left) and the GRITSBot (right).

The control gain in (18) was set to $\lambda_1 = I_5 \otimes \mathrm{diag}(2, 2, 10)$ where $I_k$ is the $k \times k$ identity matrix.

Snapshots of the initial and final formations are given in Figure 5 showing that the desired formation was successfully acquired from a random initial configuration. The path of each robot as they maneuver in formation is shown in Figure 6. Figure 7 shows the norm of all tracking and coordinations errors quickly converging to approximately zero. The errors are not exactly zero due to measurement noise and the sensor resolution. We can see from the errors that the desired pentagon formation is acquired after approximately 10 s while simultaneously. The control inputs are depicted in Figure 8, where one can see that $\eta_i(t) \to \eta_{di}$ as $t \to \infty$ for all $i$ as expected. A video of the experiment is provided as a supplementary document.

## 7.2 Adaptive Dynamics Control Simulation

Since Robotarium does not allow the specification of actuator-level commands, a simulation conducted in MATLAB was used to demonstrate the performance

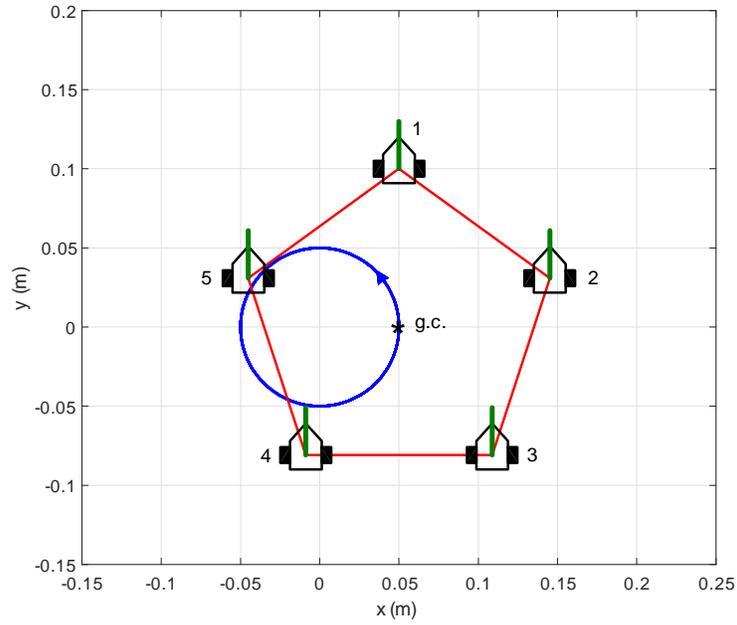

Figure 3: Desired pentagon formation at $t = 0$ along with desired circular trajectory for the geometric center.

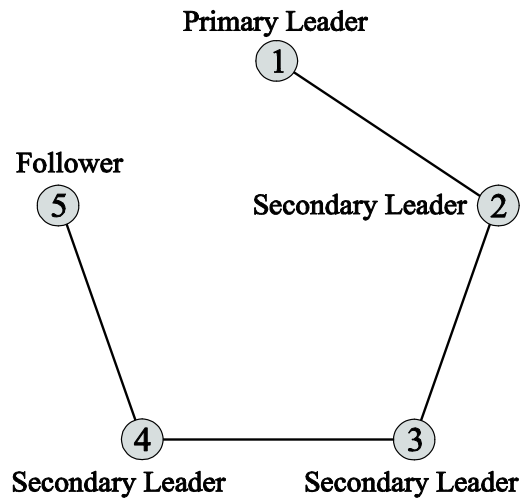

Figure 4: Graph for the pentagon formation.

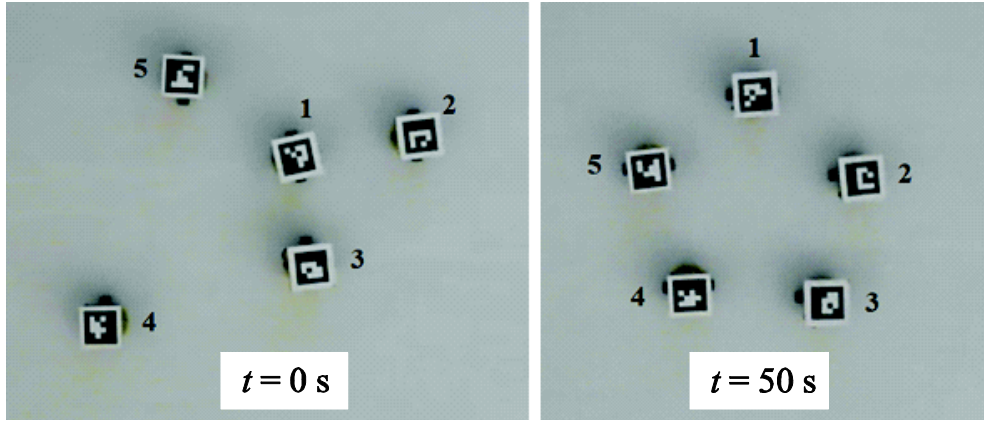

Figure 5: Snapshots of the initial formation (left) and the formation when the experiment was stopped (right).

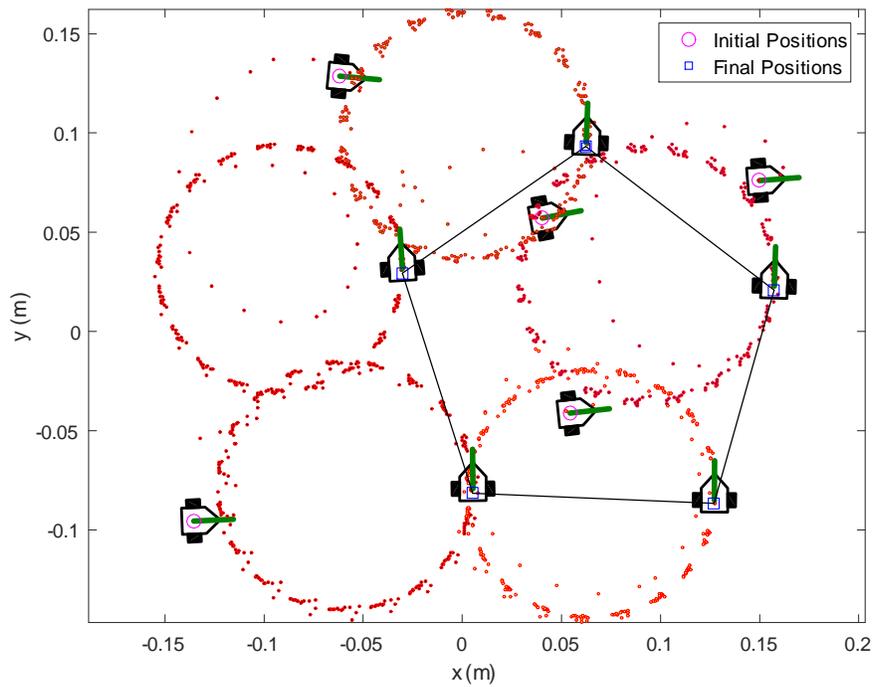

Figure 6: Circular maneuver of each robot from the initial formation at $t = 0$ s to the formation at $t = 50$ s.

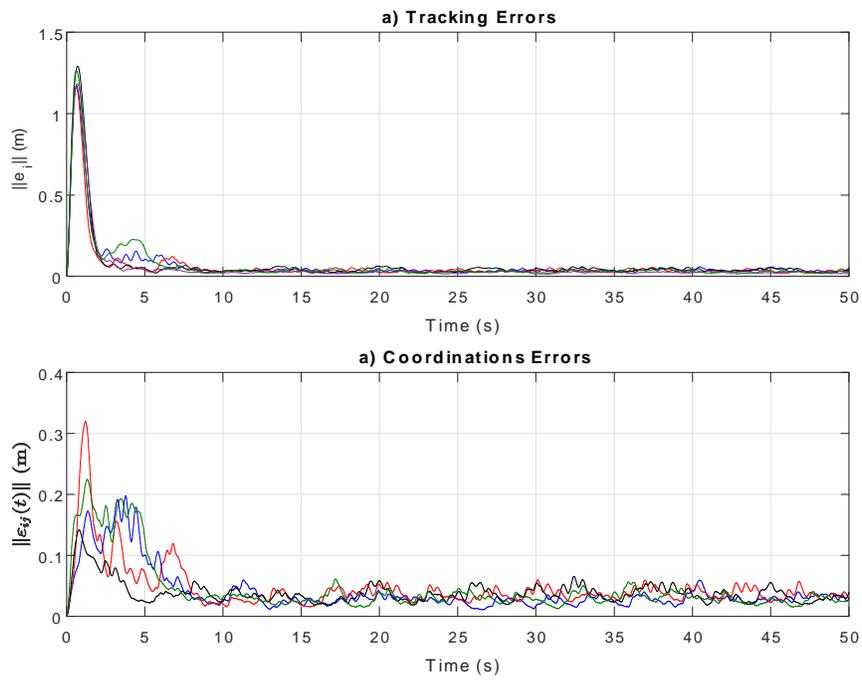

Figure 7: a) Norm of the tracking errors, $\|e_i(t)\|$, $i = 1, .., 5$; b) norm of the coordination errors, $\|\varepsilon_{ij}(t)\|$, $(i, j) \in E$.

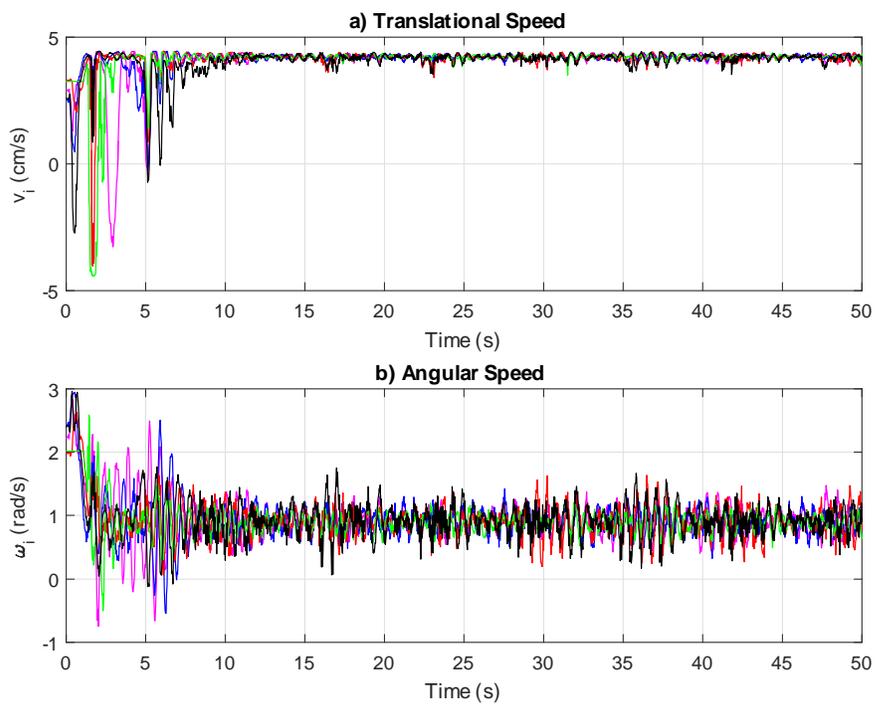

Figure 8: Control input $\eta_i(t) = (v_i(t), \omega_i(t))$, $i = 1, \dots, 5$.

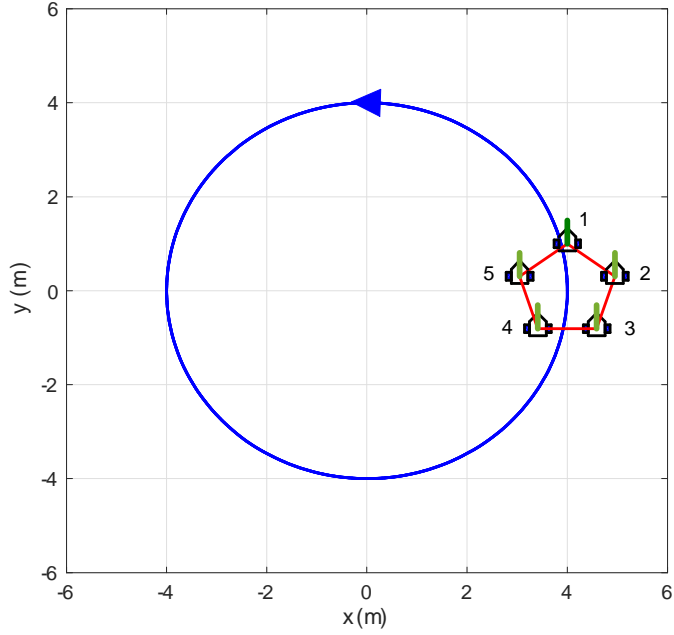

Figure 9: Desired pentagon formation at $t = 0$ along with the desired circular trajectory for the geometric center.

of the adaptive dynamics controller from Section 6. The system parameters in (1b) were set to $m_i = 3.6$ kg, $J_i = 0.0405$ kg-m$^2$, and $D_i = \text{diag}(0.3$ kg/s, $0.004$ kg-m$^2/s)$ for $i = 1, ..., 5$. The desired formation maneuver was similar to the one in the experiment with exception of $\eta_{di} = (4$ m/s, $1$ rad/s$)$, $i = 1, ..., 5$ and the initial conditions

$$q_{d1}(0) = (4 \text{ m}, 1 \text{ m}, \pi/2 \text{ rad}), \quad q_{d2}(0) = (4 + \cos \pi/10 \text{ m}, \sin \pi/10 \text{ m}, \pi/2 \text{ rad}),$$
$$q_{d3}(0) = (4 + \sin \pi/5 \text{ m}, -\cos \pi/5 \text{ m}, \pi/2 \text{ rad}), \quad q_{d4}(0) = (4 - \sin \pi/5 \text{ m}, -\cos \pi/5 \text{ m}, \pi/2 \text{ rad}),$$
$$q_{d5}(0) = (4 - \cos \pi/10 \text{ m}, \sin \pi/10 \text{ m}, \pi/2 \text{ rad}).$$

The graph was set to the one in Figure 4. The desired formation at $t = 0$ along with the desired trajectory for the geometric center of the pentagon is shown in the Figure 9.The initial conditions for the robots' position and orientation were randomly chosen as

$$q_1(0) = (0.3200 \text{ m}, 2.8857 \text{ m}, 0.0139 \text{ rad}), \quad q_2(0) = (2.3247 \text{ m}, 2.4519 \text{ m}, 2.6061 \text{ rad}),$$
$$q_3(0) = (0.2533 \text{ m}, 1.1993 \text{ m}, 0.7796 \text{ rad}), \quad q_4(0) = (2.4002 \text{ m}, 1.2942 \text{ m}, 2.7319 \text{ rad}),$$
$$q_5(0) = (0.5455 \text{ m}, 0.7914 \text{ m}, 0.4366 \text{ rad}).$$

The initial translational and angular speed of each robot was set to zero. All parameter estimates in (32) were initialized to zero, i.e., $\hat{\phi}(0) = 0$. The control and adaptation gains were set to $\lambda_1 = I_{15}$, $\lambda_2 = 3I_{10}$, and $\Gamma = I_{30}$.

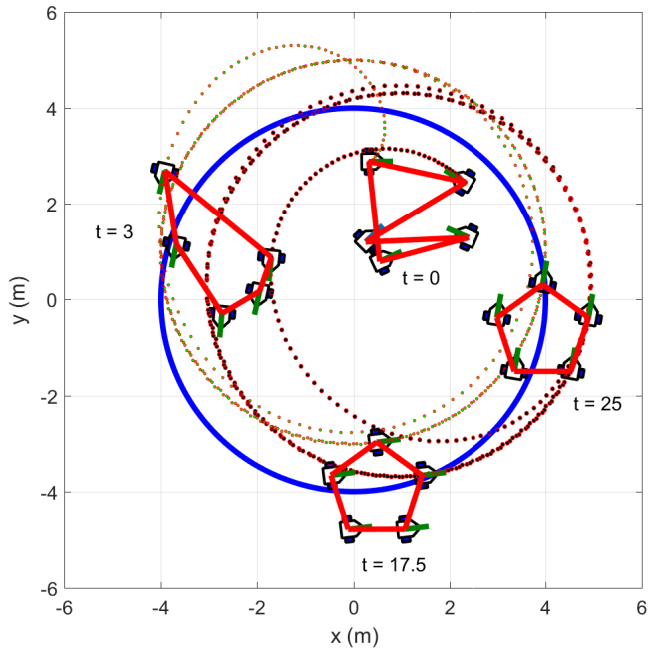

Figure 10: Snapshots in time of the formation maneuvering and the desired trajectory for the geomteric center.

Figure 10 shows the snapshots in time of the unicycle robots as they maneuver in formation. Only the trajectories of two robots are shown for illustration purposes. An animation of the formation maneuvering simulation is available in the `gif` file of the supplementary documents. Figures 11 and 12 show the time evolution of $\|e_i(t)\|$, $i = 1, \ldots, 5$ and $\|\varepsilon_{ij}(t)\|$, $(i, j) \in E$, respectively. One can see from these figures that after approximately 15 s the robots converge to the pentagon formation while maneuvering as a rigid body around the circle. The parameter estimates for robot 1 are shown in Figure 13. The parameter estimates for the other robots had a similar behavior, converging to some constant values.

## 8 Conclusion

This paper contributed a solution to the formation maneuvering problem for systems of nonholonomic, unicycle robots. The key properties of the result were the use of a spanning tree to model the inter-robot coordination graph, the use of a leader-follower type strategy motivated by the spanning tree, and the judicious use of tracking errors and coordination errors to quantify the formation maneuvering control objective and decentralize the control system. A Lyapunov-based control law was designed that ensured that the formation maneuvering task is achieved in the least squares sense. Backstepping was used to extend the kinematic control to the dynamic case with parametric uncertainty. The kinematic control law was successfully implemented on the Robotarium testbed while a simulation demonstrated the adaptive dynamics-level controller.

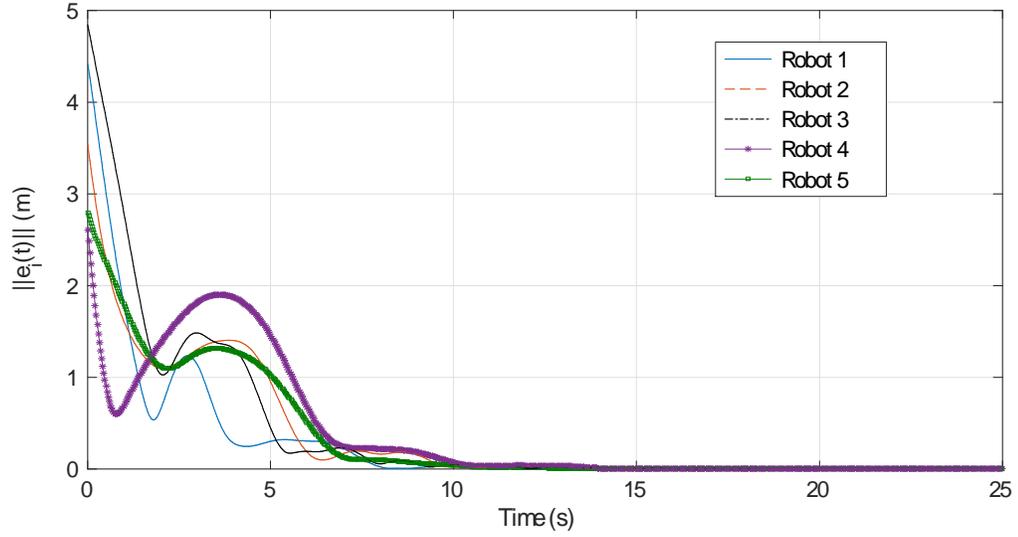

Figure 11: Norm of tracking errors, $\|e_i(t)\|$, $i = 1, .., 5$.

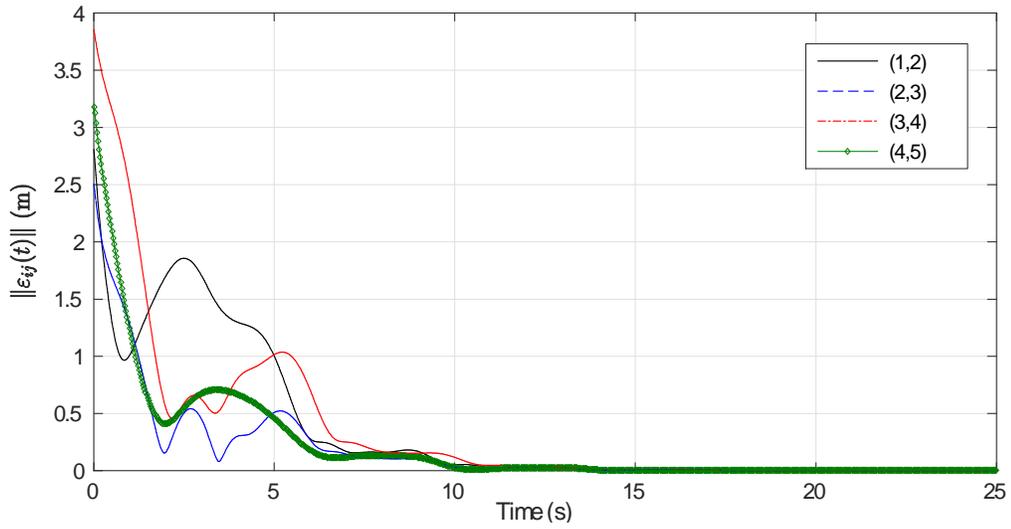

Figure 12: Norm of the coordination errors, $\|\varepsilon_{ij}(t)\|$, $(i, j) \in E$.

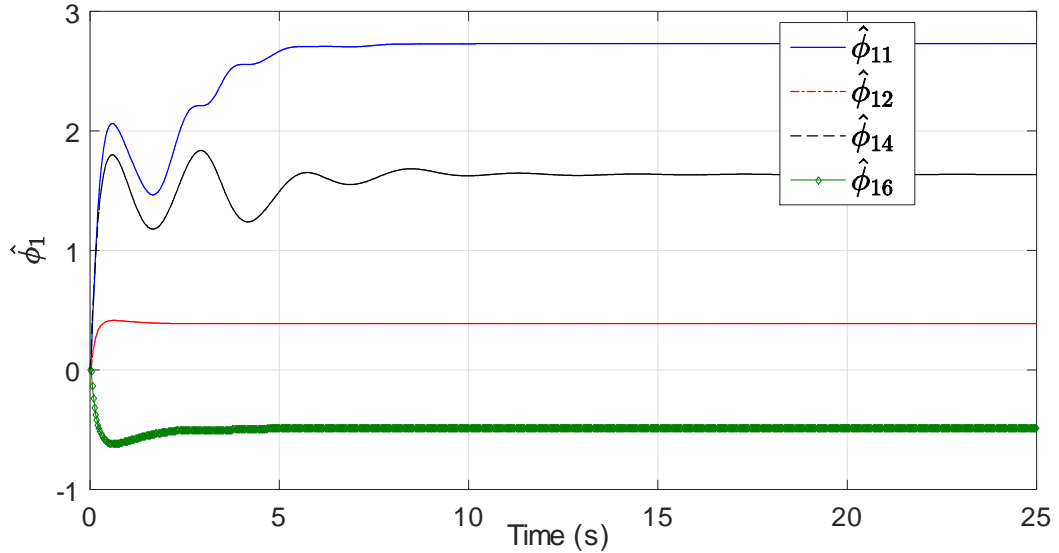

Figure 13: Sample of parameters estimates for agent 1, $\hat{\phi}_1(t)$.

The proposed formation controller does not include a collision avoidance strategy. However, one can incorporate the collision avoidance method from (22) since it only modifies the desired trajectories of robots at risk of colliding without interfering with the stability properties of the control.

# A   Appendix A

Here, we prove that $K$ in (16) has full column rank by showing that $K^\mathsf{T} K \in \mathbb{R}^{2n \times 2n}$ is invertible. First, from (2) and (11), we have that $P^\mathsf{T} P = S^\mathsf{T}(\theta_i) S(\theta_i) = I_2$ and

$$S^\mathsf{T}(\theta_i) S(\theta_j) = S^\mathsf{T}(\theta_j) S(\theta_i) = \begin{bmatrix} \cos \theta_{ij} & 0 \\ 0 & 1 \end{bmatrix} \qquad (35)$$

where $\theta_{ij} = \theta_i - \theta_j$. Using the above properties and the shorthand notations $S_i := S(\theta_i)$, we have that

$$K^\mathsf{T} K = \begin{bmatrix} 2I_2 & -S_1^\mathsf{T} S_2 & 0_{2\times 2} & 0_{2\times 2} & 0_{2\times 2} & \ldots & \ldots & 0_{2\times 2} \\ -S_2^\mathsf{T} S_1 & 2I_2 & -S_2^\mathsf{T} S_3 & 0_2 & 0_{2\times 2} & \ldots & \ldots & 0_{2\times 2} \\ 0_{2\times 2} & -S_3^\mathsf{T} S_2 & 2I_2 & -S_3^\mathsf{T} S_4 & 0_{2\times 2} & \ldots & \ldots & 0_{2\times 2} \\ \vdots & \ddots & \ddots & \ddots & \ddots & \ddots & & \vdots \\ 0_{2\times 2} & \ldots & \ldots & \ldots & 0_{2\times 2} & -S_{m-1}^\mathsf{T} S_{m-2} & 2I_2 & -S_{m-1}^\mathsf{T} S_m \\ 0_{2\times 2} & \ldots & \ldots & \ldots & & 0_{2\times 2} & -S_m^\mathsf{T} S_{m-1} & I_2 \end{bmatrix}. \tag{36}$$

For illustration purposes, for the case of 3 robots, the above matrix becomes

$$K^\mathsf{T} K = \begin{bmatrix} 2 & 0 & -\cos\theta_{12} & 0 & 0 & 0 \\ 0 & 2 & 0 & -1 & 0 & 0 \\ -\cos\theta_{12} & 0 & 2 & 0 & -\cos\theta_{23} & 0 \\ 0 & -1 & 0 & 2 & 0 & -1 \\ 0 & 0 & -\cos\theta_{23} & 0 & 1 & 0 \\ 0 & 0 & 0 & -1 & 0 & 1 \end{bmatrix} \tag{37}$$

where (35) was used.

Interestingly, $K^\mathsf{T} K$ is a *pentadiagonal* matrix (2). Specifically, an $m \times m$ pentadiagonal matrix $\Delta$ has the form

$$\Delta = \begin{bmatrix} a_1 & b_1 & c_1 & 0 & \ldots & \ldots & \ldots & 0 \\ d_2 & a_2 & b_2 & c_2 & 0 & \ldots & \ldots & 0 \\ e_3 & d_3 & a_3 & b_3 & c_3 & 0 & \ldots & 0 \\ 0 & e_4 & d_4 & a_4 & b_4 & c_4 & 0 & \ldots \\ \vdots & \ddots & \ddots & \ddots & \ddots & \ddots & \ddots & \ddots \\ 0 & \ldots & 0 & e_{m-2} & d_{m-2} & a_{m-2} & b_{m-2} & c_{m-2} \\ 0 & \ldots & \ldots & 0 & e_{m-1} & d_{m-1} & a_{m-1} & b_{m-1} \\ 0 & \ldots & \ldots & \ldots & 0 & e_m & d_m & a_m \end{bmatrix} \tag{38}$$

where $\Delta = [\gamma_{ij}]$ is such that $\gamma_{ij} = 0$ for $|i - j| > 2$.

To find the determinant of $\Delta$, we will utilize the algorithm from (41): $|\Delta| = \prod_{i=1}^m x_i$ if $x_i \neq 0$, $i = 1, \ldots, m$ where

$$x_i = \begin{cases} a_1, & i = 1 \\ a_2 - y_1 z_2, & i = 2 \\ a_i - y_{i-1} z_i - \dfrac{e_i c_{i-2}}{x_{i-2}}, & i = 3, \ldots, m, \end{cases} \tag{39}$$

$$y_i = \begin{cases} b_1, & i = 1 \\ b_i - z_i c_{i-1}, & i = 2, \ldots, m-1, \end{cases} \tag{40}$$

and

$$z_i = \begin{cases} \dfrac{d_2}{x_1}, & i = 2 \\[3mm] \dfrac{d_i - \frac{e_i y_{i-2}}{x_{i-2}}}{x_{i-1}}, & i = 3, \dots, m. \end{cases} \tag{41}$$

Note that (36) satisfies (38) with

$$\begin{aligned} &a_1 = \dots = a_{m-2} = 2, \quad a_{m-1} = a_m = 1 \\ &b_1 = \dots = b_{m-1} = d_2 = \dots = d_m = 0 \\ &c_i = \begin{cases} -\cos\theta_{\left(\frac{i-1}{2}\right)\left(\frac{i+1}{2}\right)}, & i = \text{odd} \\ -1, & i = \text{even} \end{cases} \\ &e_i = c_{i-2}, \quad i = 3, \dots, m \end{aligned} \tag{42}$$

where $m = 2n$, which is obviously even. After substituting (42) into (39)-(41), we obtain the following simplified recursive algorithm

$$x_i = \begin{cases} 2, & i = 1, 2 \\ a_i - \dfrac{e_i^2}{x_{i-2}}, & i = 3, ..., m. \end{cases} \tag{43}$$

where

$$e_i^2 = \begin{cases} \cos^2\theta_{\left(\frac{i-1}{2}\right)\left(\frac{i+1}{2}\right)}, & i = \text{odd} \\ 1, & i = \text{even}. \end{cases} \tag{44}$$

After some simple calculations, it follows from (42)-(44) that

$$\begin{aligned} x_i &= \begin{cases} 2, & i = 1, 2 \\ 2 - \dfrac{\cos^2\theta_{\left(\frac{i-1}{2}\right)\left(\frac{i+1}{2}\right)}}{x_{i-2}}, & i = 3, 5, \dots, m-3 \\ 1 + \dfrac{2}{i}, & i = 4, 6, \dots, m-2 \end{cases} \\ x_{m-1} &= 1 - \dfrac{\cos^2\theta_{\left(\frac{m-2}{2}\right)\left(\frac{m}{2}\right)}}{x_{m-3}} \\ x_m &= \dfrac{2}{m}. \end{aligned} \tag{45}$$

From (45), it is clear that $x_i \neq 0$ for $i = $ even. Furthermore, we can obtain

$$\begin{aligned} &\dfrac{i+3}{i+1} \le x_i \le 2, \quad i = 1, 3, \dots, m-3 \\ &\dfrac{2}{m} \le x_{m-1} \le 1, \end{aligned} \tag{46}$$

which indicates that $x_i \neq 0$ for $i = $ odd. Since all $x_i \neq 0$, then $|K^\mathsf{T} K| = \prod\limits_{i=1}^{m} x_i \neq 0$ and $K^\mathsf{T} K$ is invertible. The proof is complete.

# B   Appendix B

In this appendix, we compute the expression for $\dot{\eta}_f$. After taking the time derivative of (31), we have

$$
\begin{aligned}
\dot{\eta}_f \;=\; & -\frac{d\left(K^{\dagger}\left(\theta\right)\right)}{dt}\left[\lambda_2 z + H(\theta_1, t)\right] \\
& -K^{\dagger}\left(\theta\right)\left[\lambda_2 \dot{z} + \dot{H}(\theta_1, t)\right].
\end{aligned}
\tag{47}
$$

From (12) and (15), it is obvious that $\dot{z} = K(\theta)\eta + H(\theta_1, t)$. Taking the time derivative of (17) yields

$$
\dot{H} = \begin{bmatrix}
\omega_1 Q R^{\intercal}(\theta_1) S(\theta_{d1})\eta_{d1} + R^{\intercal}(\theta_1)[\omega_{d1} Q^{\intercal} S(\theta_{d1})\eta_{d1} + S(\theta_{d1})\dot{\eta}_{d1}] \\
\omega_{d1} Q^{\intercal} S(\theta_{d1})\eta_{d1} + S(\theta_{d1})\dot{\eta}_{d1} - \omega_{d2} Q^{\intercal} S(\theta_{d2})\eta_{d2} + S(\theta_{d2})\dot{\eta}_{d2} \\
\vdots \\
\omega_{d(n-1)} Q^{\intercal} S(\theta_{d(n-1)})\eta_{d(n-1)} + S(\theta_{d(n-1)})\dot{\eta}_{d(n-1)} - \omega_{dn} Q^{\intercal} S(\theta_{d_n})\eta_{dn} + S(\theta_{d_n})\dot{\eta}_{dn}
\end{bmatrix}
$$

where (9) and (11) were used, and $\dot{\eta}_{d_i}$ is the derivative of the reference signal defined in (7).

The derivative of (19) is given by

$$
\frac{d(K^{\dagger}\left(\theta\right))}{dt} = \frac{d\left(K^{\intercal}K\right)^{-1}}{dt} K^{\intercal}(\theta) + \left(K^{\intercal}K\right)^{-1}\dot{K}^{\intercal}\left(\theta\right)
\tag{48}
$$

where from (16)

$$
\dot{K}\left(\theta\right) = \Lambda\Upsilon,
\tag{49}
$$

$$
\Lambda = \begin{bmatrix}
0_{3\times3} & 0_{3\times3} & 0_{3\times3} & 0_{3\times3} & \cdots & 0_{3\times3} \\
0_{3\times3} & Q^{\intercal} & 0_{3\times3} & 0_{3\times3} & \cdots & 0_{3\times3} \\
0_{3\times3} & 0_{3\times3} & Q^{\intercal} & 0_{3\times3} & \cdots & 0_{3\times3} \\
\ddots & \ddots & \ddots & \ddots & \ddots & \ddots \\
0_{3\times3} & \cdots & \cdots & \cdots & 0_{3\times3} & Q^{\intercal}
\end{bmatrix} \in \mathbb{R}^{3n\times3n},
\tag{50}
$$

$$
\Upsilon = \begin{bmatrix}
0_{3\times2} & 0_{3\times2} & 0_{3\times2} & 0_{3\times2} & \cdots & 0_{3\times2} \\
-\omega_1 S(\theta_1) & \omega_2 S(\theta_2) & 0_{3\times2} & 0_{3\times2} & \cdots & 0_{3\times2} \\
0_{3\times2} & -\omega_2 S(\theta_2) & \omega_3 S(\theta_3) & 0_{3\times2} & \cdots & 0_{3\times2} \\
0_{3\times2} & 0_{3\times2} & -\omega_3 S(\theta_3) & \omega_4 S(\theta_4) & 0_{3\times2} & 0_{3\times2} \\
\ddots & \ddots & \ddots & \ddots & \ddots & \ddots \\
0_{3\times2} & \cdots & \cdots & 0_{3\times2} & -\omega_{n-1} S(\theta_{n-1}) & \omega_n S(\theta_n)
\end{bmatrix} \in \mathbb{R}^{3n\times2n},
\tag{51}
$$

and $Q$ was defined in (11). Since for any invertible matrix $A$,

$$
\frac{dA^{-1}}{dt} = -A^{-1}\dot{A}A^{-1},
\tag{52}
$$

we have that

$$\frac{d\left(K^\intercal K\right)^{-1}}{dt} = -(K^\intercal K)^{-1}\left[\dot{K}^\intercal\left(\theta\right)K(\theta)+K^\intercal(\theta)\dot{K}\left(\theta\right)\right]$$
$$\times(K^\intercal K)^{-1}. \tag{53}$$

From the above equations, we can see that the expression for $\dot{\eta}_f$ is dependent on the following variables: $\theta$ (through trigonometric functions), $\eta$, $z$, and $t$.